\documentclass[sigconf]{acmart}

\AtBeginDocument{%
  \providecommand\BibTeX{{%
    \normalfont B\kern-0.5em{\scshape i\kern-0.25em b}\kern-0.8em\TeX}}}

\copyrightyear{2021}
\acmYear{2021}
\setcopyright{acmcopyright}
\acmConference[SIGIR '21] {Proceedings of the 44th International ACM SIGIR Conference on Research and Development in Information Retrieval}{July 11--15, 2021}{Virtual Event, Canada.}
\acmBooktitle{Proceedings of the 44th International ACM SIGIR Conference on Research and Development in Information Retrieval (SIGIR '21), July 11--15, 2021, Virtual Event, Canada}
\acmPrice{15.00}
\acmISBN{978-1-4503-8037-9/21/07}
\acmDOI{10.1145/3404835.3462961}

\usepackage{balance}
\usepackage{subfiles} % Best loaded last in the preamble

\usepackage[ruled, lined, linesnumbered, commentsnumbered, longend]{algorithm2e}
\usepackage{xcolor}

\newcommand{\cut}[1]{}
\usepackage{bm}      % microtypography
\usepackage{amsmath}
\usepackage{todonotes}
\usepackage{mathtools}

\usepackage{graphics}
\usepackage{multirow}
\usepackage{tablefootnote}
\usepackage{enumitem}
\usepackage[font=scriptsize]{caption}

\usepackage{tabularx}
\newcommand{\red}[1]{\textcolor{red}{#1}}

\newcommand{\MJC}[1]{{\color{blue} [Mark: #1]}}
\renewcommand{\thefootnote}{\ifcase\value{footnote}\or \or$*$\or
(**)\or(***)\or(****)\or(\#)\or(\#\#)\or(\#\#\#)\or(\#\#\#\#)\or($\infty$)\fi}
\begin{document}

\fancyhead{}
%%
%% end of the preamble, start of the body of the document source.

\title{TIE: A Framework for Embedding-based Incremental Temporal Knowledge Graph Completion}

% \author{Jiapeng Wu}
% \email{jiapeng.wu@mail.mcgill.com}
% \affiliation{%

\author{Jiapeng Wu{$^{\dagger}$}}
% \authornote{work here \label{the_one}}
\affiliation{%
  \institution{McGill University, MILA}
  \city{Montreal}
  \country{Canada}}
\email{jiapeng.wu@mail.mcgill.com}

\author{Yishi Xu{$^{\dagger}$}}
\affiliation{%
  \institution{University of Montreal, MILA}
  \city{Montreal}
  \country{Canada}}
\email{yishi.xu@umontreal.ca}

\author{Yingxue Zhang}
\affiliation{%
  \institution{Montreal Research Center, Huawei Noah's Ark Lab}
  \city{Montreal}
  \country{Canada}}
\email{yingxue.zhang@huawei.com}

\author{Chen Ma{$^{\dagger}$}}
\affiliation{%
  \institution{McGill University}
  \city{Montreal}
  \country{Canada}}
\email{chen.ma2@mail.mcgill.ca}

\author{Mark Coates}
\affiliation{%
  \institution{McGill University}
  \city{Montreal}
  \country{Canada}}
\email{	mark.coates@mcgill.ca}

\author{Jackie Chi Kit Cheung}
\affiliation{%
  \institution{McGill University, MILA}
  \city{Montreal}
  \country{Canada}}
\email{jcheung@cs.mcgill.ca}

\renewcommand{\shortauthors}{Wu et al.}


\begin{abstract}
Reasoning in a temporal knowledge graph (TKG) is a critical task for information retrieval and semantic search. It is particularly challenging when the TKG is updated frequently. The model has to adapt to changes in the TKG for efficient training and inference while preserving its performance on historical knowledge. Recent work approaches TKG completion (TKGC) by augmenting the encoder-decoder framework with a time-aware encoding function. However, naively fine-tuning the model at every time step using these methods does not address the problems of 1) catastrophic forgetting, 2) the model's inability to identify the change of facts  (e.g., the change of the political affiliation and end of a marriage), and 3) the lack of training efficiency. 
To address these challenges, we present the \textbf{T}ime-aware \textbf{I}ncremental \textbf{E}mbedding (TIE) framework, which combines TKG representation learning, experience replay, and temporal regularization.
We introduce a set of metrics that characterizes the intransigence of the model and propose a constraint that associates the deleted facts with negative labels.

Experimental$^1$ results on Wikidata12k and YAGO11k datasets demonstrate that the proposed TIE framework reduces training time by about ten times and improves on the proposed metrics compared to vanilla full-batch training. It comes without a significant loss in performance for any traditional measures. Extensive ablation studies reveal performance trade-offs among different evaluation metrics, which is essential for decision-making around real-world TKG applications. 

\footnotetext{$^\dagger$Work done as an intern at Huawei Noah’s Ark Lab Montreal Research Center.}
\footnotetext{$^1$Code and data are available at: \url{https://github.com/JiapengWu/Time-Aware-Incremental-Embedding}} 
\end{abstract}

\maketitle

\section{Introduction}
Knowledge graphs (KGs), consisting of triples in the form of \emph{(head entity, relationship, tail entity)}, are effective data structures for representing factual knowledge and lie at the core of many downstream tasks; e.g.,, question answering~\cite{zhang2018variational,lukovnikov2017neural,huang2019knowledge} and web search~\cite{paulheim2017knowledge}. Although KGs enable powerful relational reasoning, they are usually incomplete. As such, inferring new facts based on existing ones in the KG, known as KG completion, is one of the most important tasks in KG research. 

Typical KGs represent knowledge facts without incorporating temporal information, which is sufficient under some circumstances~\cite{bordes2013translating, yang2014embedding, trouillon2016complex}. By additionally associating each triple with a timestamp, such as \emph{(Obama, visit, China, 2014)}, temporal knowledge graphs (TKGs) are able to consider the temporal dynamics. Usually, TKGs are assumed to consist of discrete timestamps~\cite{jiang2016towards}. They can be represented as a sequence of static KG snapshots. The task of inferring missing facts across these snapshots is referred to as temporal knowledge graph completion (TKGC).

% summary the previous work
To tackle the TKGC task, two avenues of work have been explored. The first line of models induces time-dependent representation with time-agnostic decoding functions to extend static KGC methods for capturing the temporal dynamics~\cite{dasgupta2018hyte,goel2020diachronic}. The second category of methods adopts spatial-temporal models, which leverage graph neural networks (GNNs) to capture the intra-graph structural information and inter-graph temporal dependencies~\cite{wu2020temp}. We argue that there are still several areas for improvement.
%The large volume of training data results in a long training time, meaning that it is impossible to achieve frequent model update and cannot include the most recent data in the training. 

% catastrophic forgetting
\textit{First}, previous methods do not explicitly formulate the incremental learning problem, where the change (addition and deletion) of historical information is incrementally available, and the model is expected to adapt to the changes while maintaining its knowledge about the historical facts. Naively, one might fine-tune the TKGC model with all available data at each new time step using gradient descent optimization. This, however, causes the model performance on the historical task to degrade quickly, a phenomenon known as \textit{catastrophic forgetting}~\cite{mccloskey1989catastrophic, xu2020graphsail}, which usually occurs because the model loses track of the key static features derived from earlier data. 
% deleted facts measures
\textit{Second}, previous methods usually only assess overall link prediction metrics such as Hits@10 and Mean Reciprocal Rank (MRR) while omitting the dynamic aspects of the TKG performance. 
There is an absence of metrics that can evaluate how well a model forgets deleted facts. For example, the quadruple \emph{(Trump, presidentOf, US, 2020)} is no longer true in 2021. Hence we would like the model to rank \emph{Biden} higher than \emph{Trump} given the query \emph{(?, presidentOf, US, 2021)}.
We argue that this is an essential measure of a model's effectiveness in modeling the temporal dynamics of TKGs.
\textit{Third}, as discussed in Section \ref{sec:problem_formulation}, previous TKGC methods ~\cite{dasgupta2018hyte,goel2020diachronic} conduct training and evaluation once across all the time steps. This does not satisfy the scalability and training efficiency requirements in real-world KG applications, where millions of entities and relations frequently update ~\cite{Vashishth2020Composition-based, ahrabian2020software}. 

% Move most of the following contents to method sections, leave here only the description about experiments 
% added facts

\paragraph{Present Work}
% We adapt two existing TKGC models to a realistic incremental scenario and propose a series of metrics to measure the performance of these TKGC models under the incremental setting.
We introduce a new task, incremental TKGC, and propose TIE, a training and evaluation framework that integrates incremental learning with TKGC. TIE combines TKG representation learning, experience replay, temporal regularization to improve model performance and alleviate catastrophic forgetting.

To measure TKGC models' ability to discern facts that were true in the past but false at present, we propose new evaluation metrics dubbed \textit{Deleted Facts Hits@10} (DF) and \textit{Reciprocal Rank Difference Measure} (RRD). To this end, we explicitly associate deleted quadruples with negative labels and integrate them into the training process, which shows improvement upon the two metrics compared to baseline methods.

Finally, we show that training using added facts significantly improves the training speed and reduces dataset size by around ten times while maintaining a similar ranking performance level compared to vanilla fine-tuning methods. 

We adapt HyTE~\cite{dasgupta2018hyte} and DE~\cite{goel2020diachronic}, two existing TKGC models, to the incremental learning task on wikidata12k and YAGO11k datasets. Experiments results demonstrate that the proposed TIE framework reduces training time by about ten times and improves some of the proposed metrics compared to the full-batch training. It comes without a significant loss in any traditional measures. Extensive ablation studies reveal the performance trade-offs among different evaluation metrics, providing insights for choosing among model variations. 

% To summarize, we make the following contributions regarding the issues identified above:

% \begin{enumerate}[leftmargin=*]
%     \item  
    
%     \item  We propose new evaluation metrics named \textit{Deleted Facts Hits@10} (DF) and \textit{Reciprocal Rank Difference Measure} (RRD) to measure TKGC models' ability to discern facts that were true in the past but false at present.
%     \item We empirically demonstrate the individual and combined effects of experience replay, temporal regularization, deleted facts training, and added facts training. Our proposed approach  deleted fact modeling mechanism can significantly 
    
%     We explicitly enforce the association between deleted quadruples and negative labels to better model the deleted facts in the current time step.
%     % better alleviate catastrophic forgetting while attaining the similar current and historical task performance as full-batch training. 
    
%     \item We show that training using added facts significantly improves the training efficiency and reduces dataset size by around ten times while maintaining the average accuracy loss within approximately 5\% compared to conventional full-batch training. 
% \end{enumerate}

\section{Related Work}

\subsection{Temporal KG Completion}
Existing TKGC methods can be broadly categorized into two lines of work. The first line uses shallow encoders with time-sensitive decoding functions to extend static KGC methods~\cite{Jiang_Liu_Ge_Sha_Chang_Li_Sui_2016,dasgupta2018hyte,goel2020diachronic,xu2019temporal}. For example,~\cite{dasgupta2018hyte} constrains entity and relation embeddings. The decoded scores of triples lie in different hyperplanes for each timestamp. The second line of methods uses spatiotemporal models, which leverage graph neural networks (GNNs) to capture intra-graph neighborhood information and temporal recurrence or attention mechanisms to capture temporal information~\cite{wu2020temp, jin2020recurrent, sankar2020dysat}. 
The third line of methods leverages temporal point processes to deal with continuous prediction in TKGs~\cite{trivedi2017know,trivedi2018dyrep,han2020graph}. However, this line of work is orthogonal to ours as their focus is the \emph{extrapolation} task in the TKG, which aims at predicting the future interactions among entities and relations.

In our work, we aim to provide an efficient incremental learning framework for TKGC. Hence we focus on the shallow embedding methods. 
\vspace{-1ex}
% since the models involved in the second line of work are more expensive in terms of time and space cost, and the improvement is marginal compared to the embedding-based method. In this work, we . 
% Concurrent work \cite{jain2020temporal} proposes a novel link prediction metric, for each query it considers the ranks of the correct entities not just at a single time step, but all time steps in which the facts is valid. 
% However, this metric does not explicitly address the aspects of model performance that derive from intransigence. We propose
% Second, we deal with datasets with discrete time steps. Furthermore, most of the existing work focuses on assigning higher rankings to unobserved facts, but we additionally propose to give a lower ranking for obsolete facts.
% In the standard TKGC task, model training is executed on the facts of all time steps jointly, and the model then evaluates its performance when predicting the missing facts. This task focuses on \emph{interpolation} in historical contexts rather than \emph{prediction}.
% On the other hand, the standard TKG \emph{extrapolation} task requires the prediction of the edges in the current and future time steps using a model trained on facts accumulated from earlier time steps. In this paper, we limit ourselves to tackle the \emph{interpolation} task in TKGC. 
\subsection{Incremental Learning}
As knowledge graphs evolve, more graph snapshots become available. However, deep learning models suffer from \textit{catastrophic forgetting} when existing models are incrementally fine-tuned according to the newly available data~\cite{kirkpatrick2017overcoming, castro2018end}. Various incremental learning techniques have been introduced to combat this issue for deep learning models. Our work is closely related to the experience replay and regularization-based methods. Experience replay, also referred to as reservoir sampling, retains an additional set of the most representative historical data. Rehearsal methods~\cite{rebuffi2017icarl, chaudhry2019continual, isele2018selective, prabhu2020gdumb} explicitly maintain a pool of historical data when training the model on new tasks. One of the earliest methods, iCarLR~\cite{rebuffi2017icarl}, sets the fixed number of samples for each task and selects samples that best approximate the feature mean of each class. 
Constrained optimization methods also belong to this category. Previous work \cite{lopez2017gradient, chaudhry2018efficient} exploits the stored samples to project the gradient of the current task's loss to a desired region. The objective is to ensure that the loss on the historical samples will decrease after training on the current task. This is equivalent to projecting the gradients of the current data to a direction that aligns with the gradients of the previous data. 
Regularization-based approaches consolidate previous knowledge by introducing regularization terms in the loss when learning on new data~\cite{kirkpatrick2017overcoming, castro2018end, yang2019adaptive, Zenke_Poole_Ganguli_2017}. 

More recent work has explored applying incremental learning techniques for training deep graph neural networks. GraphSAIL~\cite{xu2020graphsail} tackles the GNN-based recommendation system's forgetting issue using knowledge distillation at both node and graph levels.  ER-GNN~\cite{zhou2020continual} proposes node importance metrics and selects the most influential nodes in the graph as reservoir data. The model is fine tuned on the new data as well as the selected nodes during the training. A more relevant work~\cite{song2018enriching} applies the regularization-based method to enrich embeddings in knowledge graphs. However, the method in~\cite{song2018enriching} focuses on data synthesized by subdividing a static knowledge graph into multiple snapshots. 

In our work, we propose an end-to-end framework combining experience replay and regularization-based methods that are specifically tailored for incrementally training TKGC tasks.

\section{Problem Setup and Formulation}
In this section, we introduce notations, specify assumptions, and describe the encoder-decoder framework for the standard TKGC~\cite{wu2020temp}. This is the foundation of our TIE framework for incremental TKGC.

\subsection{Problem Formulation}\label{sec:problem_formulation}
A TKG is a sequence of KG snapshots: $\mathcal{G} = \{G^{1}, G^{2}, ..., G^{T}\}$, where $T$ denotes the total number of time steps in the TKG and  $G^{t}$ is the KG snapshot at time step $t$. Each graph is represented as a triple, i.e., $G^{t} = (E^{t}, R^{t}, D^{t})$. 
Here, $D^{t}$ denotes the set of all \emph{observed} quadruples $(s, r, o, t)$ occurring at time $t$; $E^{t}$ and $R^{t}$ denote the sets of entities and relations that are involved in at least one fact in $D^{t}$. Each quadruple contains the subject $s$, the relation $r$, the object $o$ and the time $t$. Let $\overline{D}^{t}$ denote the set of \emph{true} quadruples at time $t$ such that $D^{t} \subseteq \overline{D}^{t}, \forall t$. 
% The set of missing facts can therefore be written as $D_{test}^{t} = \overline{D}^{t} \setminus D^{t}$. 
The set of missing facts can therefore be written as $D_{test}^{t} = \overline{D}^{t} \setminus D^{t}$. 

For a quadruple $(s, r, o, t) \in D_{test}^{t}$ and its related object query $(s, r, ?, t)$, the goal of TKGC is ranking $o$ as high as possible. The goal of answering a subject query $(?, r, o, t)$ is similarly defined. 

Standard TKGC and incremental TKGC differ in terms of 1) the scope of the input, 2) the scope of the time steps targeted for evaluation, and 3) the set of candidate entities on which the score function is applied to produce the final ranking. 

\paragraph{Standard TKGC} In this setting, both training and evaluation are conducted once over time steps $1$ to $T$. During training, the model takes $D^{1}, D^{2}, \dots, D^{T}$ as input and simultaneously answers queries in each of $D_{test}^{1}, D_{test}^{2} , \dots, D_{test}^{T}$.
The set of candidate entities are those present from the beginning to the end, i.e., $E = \bigcup\limits_{i=1}^{T} E^{i}$. 

\paragraph{Incremental TKGC}
Under this setting, both training and evaluation are conducted at each time step upon the available new data $D^t$.  Hence, the input is the sequence $D^{1}, D^{2}, \dots, D^{t}$ and the goal is to answer queries in $D_{test}^{1}, D_{test}^{2} , \dots, D_{test}^{t}$.
As opposed to standard TKGC, the model only has access to entities present from the beginning to the current time step $t$, i.e., $E_{known}^{t} = \bigcup\limits_{i=1}^{t} E^{i}$

\subsection{Encoder-Decoder Framework}\label{sec:encoder_decoder}

TeMP~\cite{wu2020temp} proposes a TKGC framework with a temporal multi-relational message passing encoder and static KG decoders. However, their main focus is the encoder's design, which combines multi-relational message passing and commonly used temporal models (RNN and transformer). 
In TIE, we instead emphasize the time-aware embedding, which is composed of single-layer embedding matrices coupled with a time-agnostic decoding function $\phi$ designed for static KGC.

% We focus on graph representation learning methods based on single-layer embedding matrix and specifically designed decoding functions $\phi$. We introduce the commonly used decoding functions in both static and temporal KGC settings.

% Let $\phi(.)$ denote the score for triple $(s, r, o)$ and quadruple $(s, r, o, t)$ in the context of static and temporal KG completion, respectively. 

\paragraph{Encoder}
% Recent work extends $\phi(s, r, o)$ to time-aware decoding function $\phi(s, r, o, t)$.

Let $\bm{E} \in \mathbb{R}^{|E| \times d}$ and $\bm{R} \in \mathbb{R}^{|R| \times d}$ denote the entity and relation embedding matrices ($d$ is the embedding dimension for both entities and relations), The static entity representations for entity $i$ and relation $r$ are defined as $\bm{z}_i = \bm{E}[i]$ and $\bm{z}_r = \bm{R}[r]$.

Temporal KG embedding models derive \textit{time-aware} representations for entities and relations, so their temporal representations at time $t$ are denoted as $\bm{z}^t_i$ and $\bm{z}^t_r$. For example, the Diachronic Embedding (DE) proposed in~\cite{goel2020diachronic} applies a time-dependent function on static entity embeddings but does not differentiate between relation embeddings in different time steps:
\begin{equation}\label{equa:DE}
  \bm{z}^t_i[n] =
    \begin{cases}
      \bm{z}_i[n]\sigma(\bm{w}_i[n]t + \bm{b}_i[n]) & if 1 \leq n \leq \gamma d \,,\\
      \boldsymbol{z}_i[n] & if \gamma d < n \leq  d \,.
    \end{cases}
\end{equation}
Here $\bm{w}_i$ and $\bm{b}_i$ are entity-specific vectors with learnable parameters. The first $\gamma d$ elements of the vector capture temporal features while the last $(1-\gamma)d$ elements capture static features. The $sine$ function $\sigma$ is used as the activation function enabling the model.

\paragraph{Decoder}
Static KG models such as TransE~\cite{bordes2013translating}, DistMult~\cite{yang2014embedding} and ComplEx~\cite{trouillon2016complex} propose \textit{time-agnostic} score functions for each triple $(s, r, o)$. We denote these score functions by ``$\text{DEC}$''.

In the temporal KG representation learning methods, \textit{time-dependent} representations are taken as input to the \textit{time-agnostic} decoding function. Let $\phi^{t}$ denote the model with the parameters at time step $t$. The score for a quadruple $(s, r, o, t)$ is defined as follows:
\begin{equation}
    \phi^t(s, r, o, t)= \text{DEC}(\bm{z}^t_{s}, \bm{z}^t_{r}, \bm{z}^t_{o}) \,.
\end{equation}

% During training, the probability of the triple being valid can be interpreted as applying the sigmoid function on $\phi(s, r, o)$. 

\paragraph{Connection to incremental TKGC}
The encoder-decoder framework can be naturally adapted to incremental learning with simple fine-tuning using $D^t$ or full-batch training using $D^1, \dots, D^t$ at each time step. In the following sections, we define key metrics and propose a set of incremental learning techniques based on the encoder-decoder framework.
% In sections \ref{sec:metrics} and \ref{sec:method}, we establish a framework that augments the TKGC encoder-decoder framework with incremental learning techniques such as experience replay and parameter regularization. 
% In addition, we propose a set of key metrics to quantify the important aspect of model performance and use them to demonstrate the effectiveness of our proposed methods.
% \subsection{Evaluation Metrics}

\section{Metrics}\label{sec:metrics}
We start by introducing commonly used evaluation metrics in standard TKGC, followed by the notions of current, historical average, and intransigence measures in the context of TKGC to quantify the different aspects of model capacity. 

\subsection{Standard TKGC Metrics}
For each quadruple $(s, r, o, t) \in D^t_{test}$, we evaluate an object query $(s, r, ?, t)$ and a subject query $(?, r, o, t)$. 
Regarding the object query, we calculate the scores for all known entities, i.e., $\phi(s, r, o', t), \forall o' \in E^t$. The ranks are obtained by sorting the scores in descending order. Thereafter this is used to compute commonly used metrics such as Mean Reciprocal Rank (MRR) and Hits@k (k is usually 1, 3, and 10). 
The Hits@k is the percentage of test facts for which the correct entity's rank is at most $k$.
For $k=10$, we have the Hits@10 metrics, defined for object queries as:
\begin{equation}\label{hits}
\frac{1}{|D^{t}_{test}|} \sum_{(s, r, o, t) \in D^{t}_{test}} I(\text{rank}(o | s, r, t) \leq 10) \,,
\end{equation}
where $I$ is the indicator function. 

\cut{
The MRR at is defined as:
\begin{equation}\label{mrr}
\frac{1}{|D^{t}_{test}|} \sum_{(s, r, o, t) \in D^{t}_{test}} \frac{1}{\text{rank}(o | s, r, t)}
\end{equation}
}

\subsection{Incremental TKGC Metrics}

Since the objective of incremental TKGC is to incorporate facts from new time steps while preserving knowledge derived from the previous ones, an incremental learning approach should be evaluated based on its performance on both the \textit{current} and \textit{historical} quadruples. Additionally, we would like them to measure a model's ability to discern changes in the validity of facts at a different point in time, e.g., change of political affiliation or end of a marriage.
% We propose the current and historical average performance measures in order to quantify the first two aspects. 
\paragraph{Current and Historical Average Measure}
Let $\alpha_{t, j}$ be the Hits@10 value specified in Equation~\eqref{hits} evaluated on $D^j_{test}$, ($j \leq t$), using the model incrementally trained after time step $t$. 
% To ensure the measure's consistency after training at different time steps, when evaluating on the $j$-th time step, the ranking function considers only the known entities from the start to time step $j$, i.e., $E_{known}^j$. 
The current performance measure ($C$) is written as $C_t = \alpha_{t, t}$. 

We adapt the Average Accuracy Measure proposed in~\cite{chaudhry2018riemannian} to the TKGC setting, replacing the accuracy with the Hits@10 measure. 
The Average Hits@10 ($A$) at time step $t$ is defined as $A_t = \frac{1}{t}\sum_{i=1}^t \alpha_{t, i}$.
The higher the value of $A_t$, the better the model in terms of historical average performance, which is an important aspect for TKGC evaluation. This, to some degree, also measures whether a model is prone to \emph{catastrophic forgetting}. A model that cannot retain past knowledge would yield a much lower $A_t$ than a model trained using all the historical data. 
% Although previous work~\cite{lopez2017gradient, chaudhry2018riemannian} proposed metrics that specifically quantify the forgetting for incremental learning, we omit them for more focused evaluation in the experiments. 
\cut{
\MJC{It's not clear why the proposed metric is better than those that have already been proposed.} \red{Average accuracy is also proposed by this work. We just omitted another metric. We could also not mention forgetting measure at all.}
}
% While this metric 
% We omit the formulation for Forgetting Measure since it is not the focus of the current work. can be similarly adapted from \cite{lopez2017gradient}.

\cut{
The Forgetting Measure (F) after training at time step $t$ is:
\begin{equation}
    F_t = \frac{1}{t-1}\sum_{i = 1} ^{t - 1} f_i^t
\end{equation}
where $f_i^t$ is the forgetting on time-step $i$, computed as:

\begin{equation}
    f_i^t = \max_{l \in \{1,...,t-1\}} \alpha_{l, i} - \alpha_{t, i}
\end{equation}
\todo[inline]{Delete the mention for forgetting measure}
}

\paragraph{Intransigence Measure}
 In the context of TKGC, we define intransigence as the inability of an algorithm to identify knowledge that was true in the past but false at present. For example, after graduating from a college, a student is no longer associated with the college. 
 
 We categorize the measure into the model's ability to 1) assign a low rank to the deleted facts and 2) rank the currently valid facts above the deleted facts.
% The current and historical measures do not explicitly capture the model's ability to assign the  with lower ranks. We refer to the lack of such ability as the intransigence of the model. 
We propose Deleted Facts Hits@10 (\emph{DF}) and Reciprocal Rank Difference (\emph{RRD}) to measure the two aspects. 
The DF is analogous to the false positive rate in the classification setting, measuring the rank of the deleted triples' current time step as their time attributes. A lower DF value suggests that a model has a better capability to exclude deleted facts from the top 10 results. 

% However, a model with very low $DF$ might also have a low $C$ value, in which case the model do worse on
The RRD is defined as the pairwise difference of reciprocal ranks between each positive quadruple in the test set and each deleted fact in the previous data. RRD implicitly focuses on the cases where the rank value of either the positive object $o$ or the negative object $o'$ is low, e.g., $\frac{1}{1} - \frac{1}{10} = 0.9$, while discounting the cases where both rank values are high, e.g. $\frac{1}{1000} - \frac{1}{1010} \approx 9.9 \times 10^{-6}$. 

We define a time window ranging from $t-\tau_d$ to $t{-}1$ to limit the scope of evaluation. For every quadruple $(s, r, o, t)$, we aim to find and then evaluate the related deleted facts from this time window. We define the DF and RRD metrics for object queries at time step $t$:
\begin{equation}\label{deleted_fact_measure}
    \text{DF}_t \triangleq \frac{1}{Z_t} \sum_{(s, r, o, t) \in D_{test}^{t}} \sum_{o' \in O'_{s, r, t}} I(\text{rank}(o' | s, r, t) \leq k),
\end{equation}
\begin{equation}\label{RRD}
    \text{RRD}_t \triangleq \frac{100}{Z_t} \sum_{(s, r, o, t) \in D_{test}^{t}} \sum_{o' \in O'_{s, r, t}} \Big(\frac{1}{\text{rank}(o | s, r, t)} -  \frac{1}{\text{rank}(o' | s, r, t)} \Big),
\end{equation}
where $O'_{s, r, t}$ is the collection of negative objects and $Z_t$ is the normalizing constant:
\[
O'_{s, r, t}  = \{ o' | \exists t'  \in \{t{-}\tau_d  ...  t{-}1\}, \exists o' \in E^{t'}_{known},  (s, r, o', t') \in  D^{t'}\},
\]
\[
Z_t = \sum_{(s, r, o, t) \in D_{test}^{t}} |O'_{s, r, t}|.
\]
In practice, the RRD values are very close to zero. Hence we multiply the RRD by a factor of 100 for better presentation. The intransigence metrics for subject queries can be defined analogously.

% The set of all \emph{true} triples within the time window.  is denoted as $\overline{B}^{(t)} = \bigcup\limits_{i=max(1, t-\tau)}^{t-1} \overline{D}^{(i)}$.

% The set of \textit{deleted} facts $\overline{N}^{(t)}_{\tau}$ that were true at least once within the time window cam be written as:
% \begin{equation}
% \overline{N}^{(t)}_{\tau} = \{ (s, r, o, t) | (s, r, o) \not\in \overline{D}^{(t')} \land (s, r, o) \not\in \overline{B}^{(t)}\}
% \end{equation}

% % The metric is an directly adaptation of Hits@k measure ranking of these deleted edges is defined as follows:
% We propose Deleted Facts Hits Measure ($DF$), a direct adaptation of the Hits@k measure to the deleted edges, which, can be defined as follows:

% \begin{equation}\label{hits}
%     DF_t = \frac{1}{|\overline{N}^{(t)}_{\tau}|} \sum_{(s, r, o, t) \in \overline{N}^{(t)}_{\tau}} I(\text{rank}(o | s, r, t) \leq k),
% \end{equation}
% where the lower values of $DF_t$ suggests that model is not fixed on memoization

\section{Proposed Framework: TIE}\label{sec:method}
% Our design of incremental learning framework is based off the encoder-decoder structure in graph representation learning. In this work, We focus on shallow encoder, i.e. single-layer embedding based models with time-aware decoding functions (cite DE, Hyte). \\
We provide an overview of TIE before describing the proposed methods in detail in the following sections.

\begin{figure*}[htb]
\centering
  \includegraphics[width=0.8\linewidth]{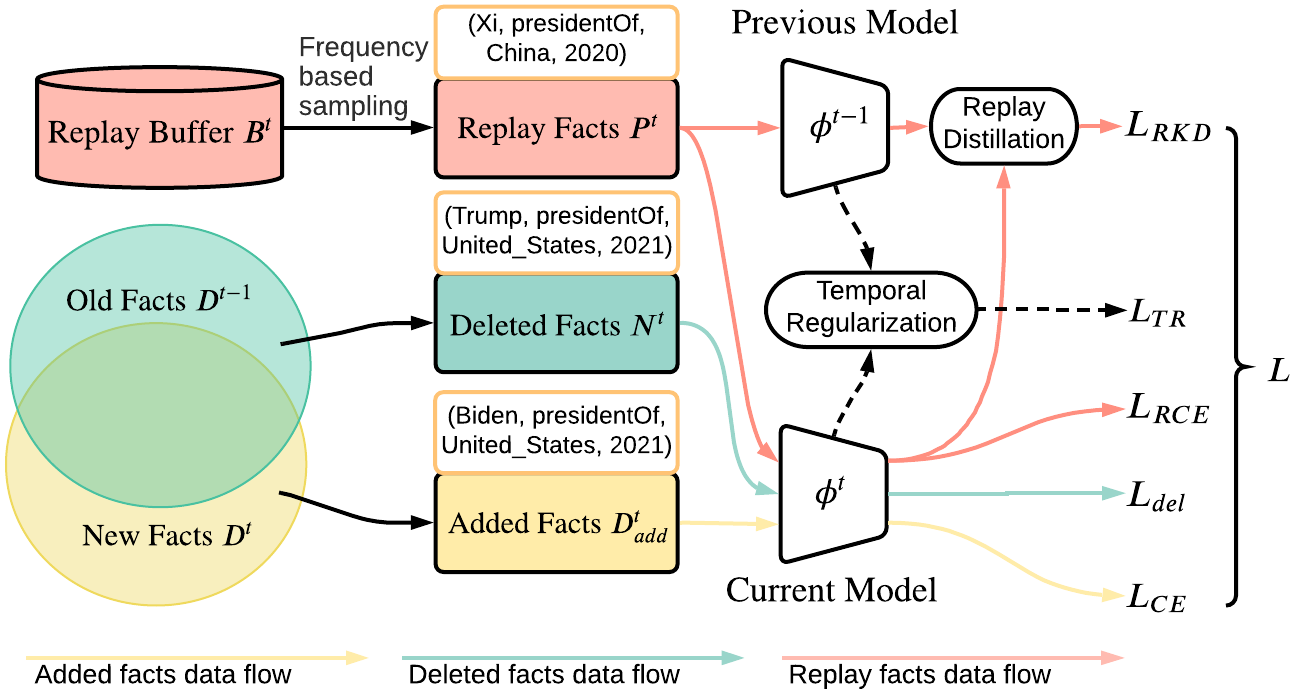}
% \vspace*{-3ex}
    \caption{A high-level illustration of the full TIE model. The four types of arrows represent the process of producing different loss terms. }
\label{fig:model_diagram}
\end{figure*}
% \vspace*{-3ex}
\subsection{Overview}
We establish the TIE framework that augments the TKGC encoder-decoder framework (Section \ref{sec:encoder_decoder}) with incremental learning techniques, a method to overcome intransigence, and an efficient training strategy. The overall architecture of TIE model is depicted in Figure \ref{fig:model_diagram}. Algorithm \ref{algorithm_core} outlines the representation learning procedure of TIE.

A key insight of our framework is that we adapt experience replay and temporal regularization techniques (Sections \ref{sec:experience_replay} and \ref{sec:regularization}) to address the catastrophic forgetting issues of fine-tuning methods using TKG representation learning models. 
Additionally, we propose to use the deleted facts from the recent time steps as a subset of negative training examples to address the \textit{intransigence} issue of the state-of-the-art TKGC methods. 
Finally, we propose to use newly added facts only for fine-tuning at each time step. This is based on the finding that the particular type of TKGs of most interest is composed primarily of persistent facts, i.e., the average duration of facts is typically long enough that no drastic changes occur between adjacent time steps. 

% using only the added facts at time $t$ compared to previous time steps
% In sections \ref{sec:experience_replay} and \ref{sec:regularization}, we propose 
% Motivated by prior work on experiment replay methods in the image classification domain (cite papers), we propose a framework that selects the historical quadruples to facilitate the learning at the current time step. 

% We first define the problem formulation for incremental TKGC and provide an overview of our framework, before describing the individual components in details. 

\subsection{Experience Replay}\label{sec:experience_replay}
% \cite{rebuffi2017icarl}
Inspired by iCaRL~\cite{rebuffi2017icarl}, we propose adapting the principle of experience replay --- to update the model parameters for the current task, we use not only the training data for the current time step but also the quadruples from earlier time steps. The data in the recent time steps can be made available with a replay buffer confined by a sliding window.

We denote the current time as $t$ and the time window length for experience replay as $\tau$. At time step $t$, we define a time window spanning from $\max(t{-}\tau, 1)$ to $t{-}1$. The historical facts in the most recent $\tau$ time steps are stored in memory, which also satisfies the resource constraints in a large-scale knowledge graph application, as it is too costly to store and load data from all time steps. 

Line 3 of Algorithm \ref{algorithm_core} constructs the replay buffer $B^t$. We first introduce two strategies for replay fact sampling, then specify the loss function, which combines a standard cross-entropy loss and a knowledge distillation loss. 
% acting as complementary input to the current data. \\ 
% We introduce the deleted facts sampling and positive facts sampling in 
\subsubsection{Replay Fact Sampling} \label{positive_sampling}
We extract $P^t$, a set of replay samples with positive labels, from $B^t$, using the following sampling strategies:

\paragraph{Uniform Sampling}
A simple yet powerful sampling strategy is to uniformly sample triples from the replay buffer $B^t$.  

\paragraph{Frequency-based Sampling}
We extend the notion of \emph{pattern frequency} introduced in~\cite{wu2020temp} to gauge the sampling probability of each quadruple in the time window and develop two approaches dubbed \emph{frequency-based sampling} and \emph{inverse frequency-based sampling}. 
A \emph{pattern} of the triple $(s, r, o)$ refers to a regular expression with some of the elements replaced with the wildcard symbol `$*$'. We use \emph{historical pattern frequency (HPF)} and \emph{current pattern frequency (CPF)} to represent the number of quadruples matching a pattern occurring before $t$ (within $B^t$) and at $t$ (within $D^t$) respectively. The set of patterns $P$ is defined as: 
\begin{equation*}\label{pattern_set}
    P = \{(s, r, o), (s, *, o), (s, r, *), (*, r, o), (s, *, o), (s, *, *), (*, *, o)\}.
\end{equation*}
Taking $(s, *, o)$ for example, the HPF $h^t_{s, *, o}$ and CPF $c^t_{s, *, o}$ are calculated as:
\begin{equation}\label{hpf}
h^t_{s, *, o} = | \{(s, r', o, t') |  \exists r', t', (s, r', o, t') \in B^t\}|,
\end{equation}
\begin{equation}\label{cpf}
c^t_{s, *, o} = | \{(s, r', o, t) | \exists r', (s, r', o, t) \in D^t\}|.
\end{equation}
The HPF and CPF for the rest of the patterns are analogously defined. 
% We use HPF and CPF to represent how many quadruples occurring \emph{before $t$} and \emph{at $t$} matches this pattern. 
The process for calculating the pattern frequencies over all quadruples and defining their sampling probabilities is highlighted in Algorithm \ref{algorithm_frequency}. 

The sampling probability for each quadruple $(s, r, o, t')$ in $B^t$ involves two separate terms: a frequency-dependent probability $fp(s, r, o)$ and a time-dependent probability $tp(t')$. We define $fp(s, r, o)$ as the weighted sum of pattern frequencies in the log scale:
\begin{equation}\label{frequency_probability}
    fp(s, r, o) = \sum_{p \in P} \lambda_p  \left[ log(h^t_p + 1) + \gamma \tau log(c^t_p + 1) \right],
\end{equation}
where the scalar value $\lambda_p$ denotes the weight associated with the frequency of pattern $p$. Recall that $\tau$ is the window length. We additionally introduce $\gamma$ as a discount factor controlling the ratio of $h^t_p$ and $c^t_p$. We take pattern frequencies in their log forms (after adding 1 to avoid zero-values) to downscale the patterns with particularly large frequency, in order to avoid repeatedly sampling quadruples with a few very frequent patterns (examples are presented later).

The term $tp(t')$ is formulated as an exponential decay function on the temporal distance to the current time step, aiming at down-weighting the quadruples from the older time steps relative to the more recent ones. It is defined as $tp(t') = \exp(\frac{t' - t}{\sigma})$, where $\sigma$ is a scalar parameter controlling the smoothness of the function. The resulting unnormalized sampling rate $s(s, r, o, t')$ is written as $fp(s, r, o)tp(t')$ (line 10, Algorithm \ref{algorithm_frequency}). 

This design encourages the sampling of quadruples with \emph{higher} pattern frequencies and discourages the sampling of quadruples with \emph{lower} pattern frequencies. For example, in the Wikidata12k dataset, \emph{(La Chapelle-sur-Oudon, instance of, commune of France, 76)} would get a high sampling probability because the pattern \emph{(*, instance of, commune of France)} has a large number of matches from time steps 62 to 77. Moreover, every quadruple with this pattern is assigned a higher sampling probability, resulting in a non-representative subset sampled by the algorithm, leading to higher intransigence.

We thus propose a simple alternative that instead encourages the sampling of quadruples with \emph{lower} pattern frequencies. We name this \emph{inverse frequency-based sampling}, and define the sampling rate as $\frac{tp(t')}{fp(s, r, o)}$ (line 8, Algorithm \ref{algorithm_frequency}). This design results in a more diverse set of replay samples, with the quadruples with high pattern frequencies discarded.

After computing the sampling rates for all the quadruples, we normalize the sampling probability for each quadruple and use them to sample $P^t$ (lines 13 -- 17, Algorithm \ref{algorithm_frequency}).

\subsubsection{Representation Learning}
For each quadruple in $P^t$, we use \textit{time-dependent negative sampling}~\cite{dasgupta2018hyte} to collect a negative set of entities to approximate the cross-entropy loss. Each negative entity is collected from the set of known entities up to time step $t'$. 
For each triple $(s, r, o, t') \in P^t$, the negative entities set is written as $D^-_{s, r, t'} = \{o' | o' \in E^{t'}_{known}, (s, r, o', t') \not\in D^{t'} \}$.

As in iCarRL, we additionally use the knowledge distillation loss~\cite{hinton2015distilling} to ensure that the previously learned discriminative information is not lost during the new current learning step. 
Before training the parameters at time step $t$, we store the output of the model with the network parameters after training at time step $t{-}1$ as:
\begin{equation}
    q^{t-1}_{s, r, o, t'} = \frac{\exp(\phi^{t-1}(s, r, o, t'))}{\sum_{o' \in D^-_{s, r, t'}} \exp(\phi^{t-1}(s, r, o', t'))}.
\end{equation}
After each training iteration, we cache the output logits $q_{s, r, o, t'}$ using the current decoding function $\phi^t$:
\begin{equation}
    q^t_{s, r, o, t'} = \frac{\exp(\phi^t(s, r, o, t'))}{\sum_{o' \in D^-_{s, r, t'}} \exp(\phi^t(s, r, o', t'))}.
\end{equation}
We use $D_{KL}$ to denote the Kullback–Leibler divergence. The replay knowledge distillation (RKD) loss and replay cross-entropy (RCE) at each iteration are defined as follows:

\begin{align}
    L_{RKD} &= \sum_{s, r, o, t' \in P^t} D_{KL}(q^{t-1}_{s, r, o, t'} || q^t_{s, r, o, t'}) \nonumber \\
           &= \sum_{s, r, o, t' \in P^t}  q^{t - 1}_{s, r, o, t'} log \big(\frac{q^{t - 1}_{s, r, o, t'}}{q^t_{s, r, o, t'}}\big),\\ \nonumber\\
    L_{RCE} &= -\sum_{s, r, o, t' \in P^t} log( q^t_{s, r, o, t'}). \label{equation:rce}
\end{align}

%\begin{equation}
%\begin{split}
%    L_{RKD} & = \sum_{s, r, o, t' \in P^t} D_{KL}(q^{t-1}_{s, r, o, t'} || q^t_{s, r, o, t'})\\
%           & = \sum_{s, r, o, t' \in P^t}  q^{t - 1}_{s, r, o, t'} log \big(\frac{q^{t - 1}_{s, r, o, t'}}{q^t_{s, r, o, t'}}\big),
%\end{split}
%\end{equation}

%\begin{equation}\label{equation:rce}
%    L_{RCE} = -\sum_{s, r, o, t' \in P^t} log( q^t_{s, r, o, t'}).
%\end{equation}

% We use both cross entropy (CE) loss and knowledge distillation (KD) loss for the sampled positive samples, and use CE loss for negative samples only.

    % \vspace{-3ex}
\begin{algorithm}
    \SetKwInOut{Input}{Input}
    \SetKwInOut{Output}{Output}
    \SetKwInOut{Require}{Require}
    
    \Input{Replay buffer $B^t$ and current facts $D^t$} 
    % \Require{Model parameter $\bm{\theta}^{t - 1}$}

    \For{$(s, r, o, t') \in B^t$}{
        Define the complete set of patterns $P$ from equation (\ref{pattern_set})
        \For{$p \in P$}{
            Calculate $h^t_{p}$ and $c^t_{p}$ by variations of equations (\ref{hpf}) and (\ref{cpf})
            % Let $e$ denote the regular expression without parenthesis corresponding to $e'$, such that $(e, t)$ represents $(s, *, o, t)$ for $e' = (s, *, o)$\\
            % $h^t_{e} = |\{e | \exists t', (e, t') \in B^t\}|$\\
            % $c^t_{e} = |\{e | (e, t) \in D^t\}|$ 
        }
        Calculate $fp(s, r, o)$ using equations (\ref{frequency_probability}) \\
        $tp(t') = \exp(\frac{t' - t}{\sigma})$\\
       \eIf{inverse frequency sampling}{
            $\psi(s, r, o, t') = \frac{tp(t')}{fp(s, r, o)}$
        }{
            $\psi(s, r, o, t') = tp(t')fp(s, r, o)$
        }
    }
    \tcc{Normalize sampling probabilities for all $(s, r, o, t') \in B^t$}
    \For{$(s, r, o, t') \in B^t$}{
    $p(s, r, o, t') = \frac{\psi(s, r, o, t')}{\sum_{\eta \in B^t}\psi(\eta)}$\\
    }
    Sample $P^t$ from $B^t$ by the normalized probability distribution
    \caption{Sampling Probability for Frequency-based Sampling at Time Step $t$}
    \label{algorithm_frequency}
\end{algorithm}

    % \vspace{-4ex}
%  algorithm
\newcommand\mycommfont[1]{{#1}}
\SetCommentSty{mycommfont}
\begin{algorithm}
    \SetKwInOut{Input}{Input}
    \SetKwInOut{Output}{Output}
    \SetKwInOut{Require}{Require}
    
    \Input{Quadruples in the current and historical time steps: $D^1, \dots, D^{t- 1}, D^t$}
    % \Output{Optimized parameters $\theta$}
    \Require{Model parameter $\bm{\theta}^{t - 1}$}
    Construct the set of added quadruples $D^t_{add}$ using Equation (\ref{equation:added_facts})\\
    Initialize model parameters $\bm{\theta}^t$ by Equation \ref{equa:initialization}\\
    Construct a replay buffer
    $B^t = \bigcup\limits_{i = \max(1, t - \tau)}^{t-1} D^{i}$\\ 
    Sample positive replay samples $P^t$ by Algorithm \ref{algorithm_frequency}\\
    Sample a set of deleted facts $N^t$ by Equation \ref{deleted_samples}\\
        
    Run network training and update model parameter with loss function (Equation \ref{equation:final_loss})\\

\caption{TIE Representation Learning at Time Step $t$}
\label{algorithm_core}
\end{algorithm}
    % \vspace{-4ex}
\subsection{Temporal Regularization}\label{sec:regularization}
Let $\bm{\theta}^{t-1}$ denote the models parameters after training at time $t-1$. We use Diachronic Embedding (DE) for demonstration, and a similar procedure can be applied to HyTE. The parameter set is $\bm{\theta}^{t-1} = \{\bm{E}^{t-1}, \bm{R}^{t-1}, \bm{W}^{t-1}, \bm{B}^{t-1}\}$, where $\bm{W}$ and $\bm{B}$ are the matrices of weight and bias parameters in Equation~\eqref{equa:DE}. 
We use $\bm{\theta}^{t-1}$ to initialize $\bm{\theta}^t$ before training at time step $t$.
The representations for entities and relations known at time step $t-1$ are initialized using the corresponding entry in $\bm{\theta}^{t-1}$, or with Glorot Initialization \cite{glorot2010understanding} if an entity appears for the first time.
The initialization for the entity embedding matrix $\bm{E}^{t}$ is as follows:
\begin{equation}\label{equa:initialization}
    \bm{E}^t_i =
    \begin{cases}
        \bm{E}^{t-1}_i , & \text{if } i \in E^{t-1}_{known}, \\
        uniform(-\sqrt{\frac{12}{d}}, \sqrt{\frac{12}{d}}), & \text{otherwise}.
    \end{cases}
\end{equation}
The initialization for other parameters are defined similarly.

Inspired by previous work~\cite{song2018enriching}, we propose temporal regularization on the parameter space to alleviate catastrophic forgetting. We impose an $L_2$ regularization constraint in the context of TKGC to smooth drastic change in the current representations compared to the previous task's parameters.

Taking the entity embedding matrix for example, we only regularize the representation for those entities that are also present in $E^{t-1}_{known}$. We use the hat symbol to denote such subsets in an embedding matrix, e.g., $\bm{\hat{E}}^t = \bm{E^t}[E^{t-1}_{known}]$.
The temporal regularization loss for DE is defined as:
\begin{multline}
    L_{TR} = ||\bm{\hat{E}}^t - \bm{E}^{t-1}||_2 + ||\bm{\hat{W}}^t - \bm{W}^{t-1}||_2 \\ + ||\bm{\hat{B}}^t - \bm{B}^{t-1}||_2 + ||\bm{\hat{R}}^t - \bm{R}^{t-1}||_2.
\end{multline}
\vspace{-3ex}
\subsection{Learning with Deleted Facts} \label{sec:deleted_facts_sampling}

To reduce the intransigence of the model as defined in Equations~\eqref{deleted_fact_measure} and~\eqref{RRD}, we propose training the model using a set of deleted quadruples from the perspective of time step $t$:
\begin{equation}\label{deleted_samples}
    N^t = \{(s, r, o, t) | (s, r, o, t) \not\in  D^{t} \land \exists t', (s, r, o, t') \in B^t \} .
\end{equation}
We associate each quadruple in $N^t$ with a negative label and calculate the binary cross entropy loss as:
\begin{equation}
    L_{del} = -\sum_{(s, r, o, t) \in N^t} log \big(1 - \sigma(\phi(s, r, o, t))),
\end{equation}
where $\sigma$ denotes the sigmoid function. We do not include knowledge distillation since the labels of $(s, r, o, t{-}1)$ and $(s, r, o, t)$ are not necessarily identical. 

It is optional to add a positive example $(s, r, o, t')$ associated with each negative example $(s, r, o, t)$ to $N^t$. However, we do not find it helpful experimentally in terms of alleviating intransigence.

\subsection{Learning with Added Facts}\label{sec:added_facts}
\begin{figure}[!tb]
\centering
  \includegraphics[width=1\linewidth]{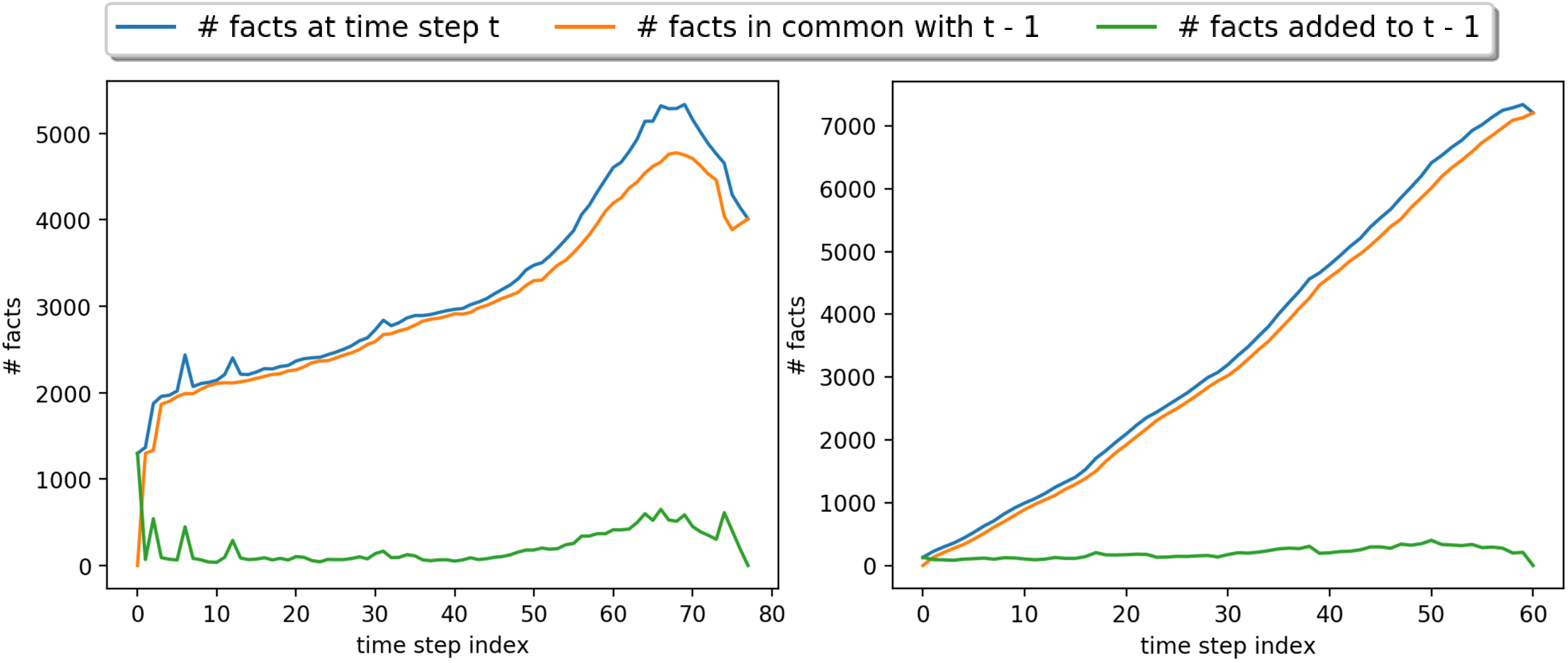}
    \caption{Dataset statistics of Wikidata12k (left) and YAGO11k(right). The three curves represent the numbers of facts (total, common and added) at every time step. }
\label{fig:dataset_stats}
\vspace{-4ex}
\end{figure}

We observe via statistics of Wikidata12k and YAGP11k (Figure \ref{fig:dataset_stats}) that most facts are common between time steps $t{-}1$ and $t$, besides the time attributes.
This suggests that fine-tuning using all the quadruples within $D^t$ essentially re-emphasizes the majority of the facts that the model has previously seen. 

We propose a novel training strategy that uses only the added facts at each time step, i.e., facts that just become true at $t$ despite being false at $t{-}1$. This can significantly reduce the size of training data, thus accelerating training by orders of magnitude. It also allows us to incorporate other complementary techniques introduced in Sections \ref{sec:experience_replay} -- \ref{sec:deleted_facts_sampling} without sacrificing training efficiency, compared to fine-tuning with $D^t$ alone. 

The added facts at time $t$ and the corresponding loss functions are defined as follows:
\begin{align}\label{equation:added_facts}
    D^t_{add} &= \{ (s, r, o, t) | (s, r, o, t) \in  D^{t} \land (s, r, o, t{-}1) \not\in D^{t{-}1} \},\\
    L_{CE} &= -\sum_{(s, r, o, t)  \in D^t_{add}} log( q^t_{s, r, o, t}).
\end{align}

%\begin{equation}\label{equation:added_facts}
%    D^t_{add} = \{ (s, r, o, t) | (s, r, o, t) \in  D^{t} \land (s, r, o, t{-}1) \not\in D^{t{-}1} \},
%\end{equation}
%\begin{equation}
%    L_{CE} = -\sum_{(s, r, o, t)  \in D^t_{add}} log( q^t_{s, r, o, t}).
%\end{equation}
% $D^{t-1}$ and $D^t$, which consists of deleted facts (valid at $t-1$ but not $t$) and added triples (valid at $t$ but not $t-1$). The treatment for deleted triples are discussed in equation \ref{deleted_samples}, we propose replacing the current training quadruples $D_t$ with added quadruples:
% The size of $D^{t}_{add}$ is usually orders of magnitudes lower that $D^t$, which could increase the training efficiency and data efficiency in the current task setting. We use standard cross-entropy loss:

\subsection{Optimization}\label{sec:optimization}
\subsubsection{Summation of loss functions}
The final loss function for TIE is defined as the weighted combination of the loss terms defined in Sections \ref{sec:experience_replay}--\ref{sec:added_facts}:
\begin{equation}\label{equation:final_loss}
L = \alpha_1 L_{CE} + \alpha_2 L_{del} + \alpha_3 L_{RCE} + \alpha_4 L_{RKD} + \alpha_5 L_{TR},
\end{equation}
where the $\alpha$'s are weight parameters controlling the relative emphasis placed on each loss term.

\cut{
\subsubsection{A-GEM Optimization}
We also investigate the adaptation of A-GEM~\cite{chaudhry2018efficient} for incremental TKGC. Please refer to the Appendix for details. 
}

\section{Experiments}

We evaluate the performance of our models on two standard TKGC benchmark datasets using our proposed evaluation protocol. We also conduct various ablation studies investigating the effectiveness of individual and combined components of the proposed methods.

\subsection{Datasets}
Common TKGs such as YAGO3~\cite{leblay2018deriving} and Wikidata~\cite{erxleben2014introducing} have valid time intervals associated with a subset of the facts. We use the two instances in these datasets, i.e., YAGO11k and Wikidata12k released in~\cite{dasgupta2018hyte} for the experiments. 
The statistics of the two datasets are summarized in Table \ref{tab:dataset_table}. 
\begin{table}[ht]
  \resizebox{\columnwidth}{!}{
  \begin{tabular}{cccccccc}
  \toprule
     Dataset & $|E|$ & $|R|$ & $T$ & |train|   & |valid| & |test| & |total| \\
     \midrule
     Wikidata12K & 12554 & 24 & 78 & 257,542 & 20,764 & 19,746 & 298,052 \\
     YAGO11K & 10623 & 10 & 61 & 215,894 & 23,197 & 22,567 & 261,658 \\
     \bottomrule
  \end{tabular}
  }
  \caption{Statistics of Wikidata12K and YAGO11K. }
  \label{tab:dataset_table}
\end{table}
\paragraph{YAGO11k} This is a subset of YAGO3~\cite{mahdisoltani2013yago3}.  In the YAGO3 dataset, the valid time of a fact is represented as a time interval, e.g., \textit{(Pétala Monteiro, isAffiliatedTo, Democratic Labour Party (Brazil), [1999-\#\#-\#\#, 2014-\#\#-\#\#])}. We follow the same preprocessing procedures described in~\cite{dasgupta2018hyte} including subgraph extraction and conversion from time interval to discrete time steps. 

We also follow the practice of merging multiple adjacent time steps to balance the number of triples in different time steps, resulting in 61 different time steps in total. 
However, an issue with the merging is that redundant facts occur within both training and validation sets. We fix the problem by retaining only one occurrence for each quadruple in the dataset.

For a time interval with a missing start or end date, e.g., \textit{(Ann Shoemaker, isMarriedTo, Henry Stephenson, [1956-\#\#-\#\#, \#\#\#\#-\#\#-\#\#])}, we use the first and the last time step in the entire dataset to represent the missing start time or end time. 

\paragraph{Wikidata12k} This is a subset of Wikidata~\cite{erxleben2014introducing}. Similar to YAGO11k, Wikidata12k associates each fact with a time interval. We create a TKG with 77 time-steps by applying the same processing as YAGO11k.

\subsection{Experiment Setting}

\subsubsection{Training and Evaluation Protocol}
\begin{figure}[ht]

\vspace*{-2ex}
\centering
  \includegraphics[width=1\linewidth]{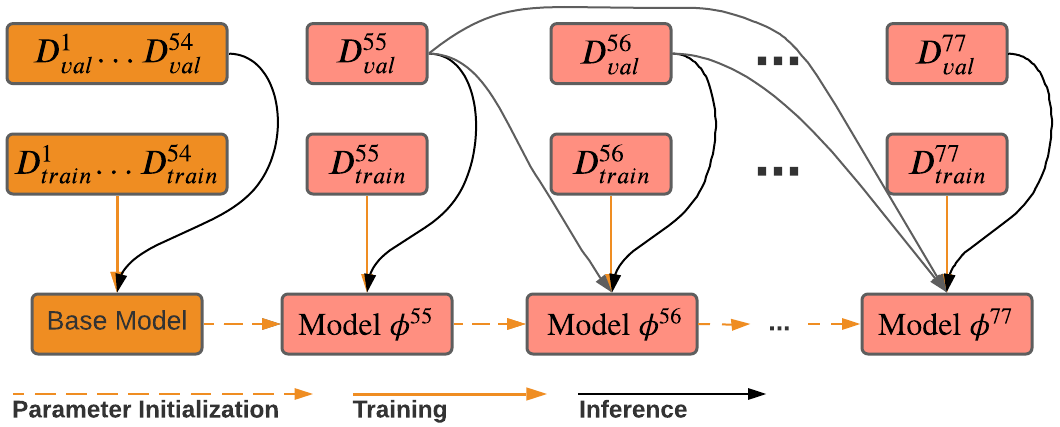}
  
\vspace*{-2ex}
    \caption{Training and evaluation diagram for Wikidata12K dataset.}
    
\vspace*{-2ex}
\label{fig:trainin_eval_figure}
\end{figure}

% base model and incremental model training
Each set of quadruples $D^t$ is partitioned into $D^t_{train}$, $D^t_{val}$ and $D^t_{test}$. 
Motivated by the experiment setting in~\cite{xu2020graphsail}, we pretrain a base model using the quadruples from the first 70\% of the time steps following standard TKGC training and evaluation protocols. We then perform incremental training and evaluation on the last 30\% of the time steps. Figure \ref{fig:trainin_eval_figure} illustrates the training and evaluation procedure for Wikidata12k.

At each time step $t$, we use the current performance measure $C$ as the criterion for early-stopping on the $D^t_{val}$ with patience set to 20. We use the best model at the current time step to compute all metrics defined in Section~\ref{sec:metrics} on the test set to obtain the final performance. For evaluation, the time window length $\tau_d$ for calculating $DF$ and $RRD$ is set to 10. 

We report results averaged across 4 randomized runs, along with standard deviation.

\subsubsection{Evaluation Metrics}
% average metrics
We report the following metrics defined in section \ref{sec:metrics}:
\begin{itemize}
    \item \textbf{C@10}, or C-Hits@10 ($\uparrow$): Current Hits@10.
    \item \textbf{DF@10}, or DF-Hits@10 ($\downarrow$) : Deleted Facts Hits@10.
    \item \textbf{RRD} ($\uparrow$) : Reciprocal Rank Difference.
    \item \textbf{A@10}, or A-Hits@10 ($\uparrow$) : Average Hits@10.
\end{itemize}
The symbol $\uparrow$ suggests the higher the metric value, the better the performance. The opposite is true for $\downarrow$.

For each evaluation metric, we calculate the average across the incremental learning time steps. For example, the reported current Hits@10 on Wikidata12k is $\frac{1}{22} \sum_{t = 55}^{77}C_t$, as shown in figure \ref{fig:trainin_eval_figure}, time step 55 is the beginning of the incremental learning.

We also report the average training time per epoch, and the size of training data at each time step.
Note that the training time includes only the GPU computation time and omits any sampling conducted on the CPU to ensure a fair comparison. Dataset size represents the total number of quadruples involved in the training, including both current facts, experience replay facts, and the sampled negative quadruples.

\subsubsection{Representation Learning Base Models}
We directly adapt \textbf{DE}~\cite{goel2020diachronic} and \textbf{HyTE}~\cite{dasgupta2018hyte} to the incremental learning setting. The exact formulations are introduced in Section \ref{sec:encoder_decoder}. 
We use ComplEx~\cite{trouillon2016complex} and TransE~\cite{bordes2013translating} as the decoding functions for DE and HyTE, respectively. 

\subsubsection{Baselines and Skylines Algorithms}
% \red{describe each one\\}
To demonstrate the effectiveness of our incremental training framework, we compare three incremental training strategies, including Fine-tune (\textbf{FT}), Temporal regularization (\textbf{TR}), and the complete proposed model \textbf{TIE}. 
\begin{enumerate}
    \item Fine-tune (\textbf{FT}): this is a naive baseline that only utilizes added facts $D^t_{add}$ and corresponding cross-entropy loss to fine-tune the model at each time step. 
    \item Temporal Regularization (\textbf{TR}): TR uses temporal regularization loss (Section \ref{sec:regularization}) in addition to FT, and it uses the same data for training.
    \item The proposed complete model (\textbf{TIE}): TIE follows all the procedures described in Algorithm $\ref{algorithm_core}$. It includes training using added facts, deleted facts, temporal regularization and experience replay with positive facts. The choice of experience replay sampling strategy (random, frequency-based or inverse-frequency based) is conditioned on the validation set performances for each experiment setting. 
\end{enumerate}

We also provide two skyline training mechanisms --- Full Batch (\textbf{FB}) and Full Batch training with future facts (\textbf{FB\_future}) to showcase the approximate accuracy upper-bound of our proposed incremental training algorithm.
\begin{enumerate}
    \item Full-batch (\textbf{FB}): In this setting, we use all the quadruples in $B^t$ and $D^t$ to fine-tune the model. Negative sampling is based on facts at different time steps. No regularization or distillation is involved.
    \item Full-batch with future data (\textbf{FB\_future}): This setting acts as an oracle with access to all data in $D^1, ..., D^T$. The training and evaluation protocol is described in Section \ref{sec:problem_formulation}. 
\end{enumerate}

\subsection{Implementation and Hyperparameters}
% training
Our models are implemented using PyTorch with the support of the PyTorch-lightning library~\cite{falcon2019pytorch}.
We set the learning rate to $10^{-3}$ and embedding sizes to 128 for all experiments. We use 2048 as the batch size. When training using both current quadruples, experience replay, and deleted facts, we ensure that the total number of these samples in each batch is at most 2048. 
For the training data at the current time step, we use 500 as a negative sampling rate, meaning that we sample 500 negative entities for each query. Because we corrupt subjects and objects separately, there are in total 1000 negative samples collected to estimate the probability of a factual triple. 
For the replay samples, we set the negative sampling rate to 50, i.e., sampling 100 negative entities for each replay fact. We find that this ratio achieves an appropriate trade-off between task performance and training time. The above hyperparameters are selected based on the validation set performances. 
For experience replay experiments, the training time window length $\tau$ is set to 10, and we sample 1,000 historical samples per time step. 

For positive reservoir sampling (described in Section \ref{sec:experience_replay}), we set $\sigma = 10$ and $\gamma = 0.5$. We list the $\lambda$ values associated with each pattern in $P$ (Equation \ref{pattern_set}):
$\lambda_{(s, r, o)} = 2, \lambda_{(s, *, o)} = 1.5, \lambda_{(s, r, *)} = \lambda_{(*, r, o)} = 1.3, \lambda_{(s, *, * )} = \lambda_{(*, *, o )} = 1$, $\lambda_{(*, r, * )} = 0$. We set all $\alpha$'s to 1, thus imposing all the loss terms with equal weights. 

% \subsection{Results and Analysis}

\subsection{Comparative Study}

\begin{table*}[htb] 

%   \resizebox{\columnwidth}{!}{
    % \small
  \begin{tabular}{c|c|ccccccc}
    \toprule
    \multicolumn{1}{c}{Dataset} & \multicolumn{1}{c}{Base model} & Algo. & C@10 ($\uparrow$) & DF@10 ($\downarrow$) & RRD ($\uparrow$) & A@10 ($\uparrow$) & Training time(s)$^1$ \tablefootnote{$^1$Training time refers to the average time needed to train the model for one epoch. To compare the efficiency fairly, we conduct experiments of different models on the same machine (with a single NVIDIA Tesla V100 GPU).} & Data size  \\
    \midrule
    \midrule
    \multirow{10}{*}{Wikidata12k} & \multirow{5}{*}{DE}    & FT    & 31.90	$\pm$ 0.42 & 12.50 $\pm$ 0.10 & -4.16 $\pm$ 0.07 & 36.05 $\pm$ 0.52  & 0.89  &  424.7K	\\
                            &                       & TR    & 34.14 $\pm$ 0.17 &	13.05 $\pm$ 0.08	& -4.45 $\pm$ 0.09 &	\textbf{38.43} $\pm$ 0.13 & 1.27  &  424.7K  \\
                            &                       & TIE & \textbf{34.90} $\pm$ 0.18	& \textbf{8.76} $\pm$ 0.67 & \textbf{-1.89} $\pm$ 0.36 	& 37.57 $\pm$ 0.07 & 15.79 &  1.41M	\\
                            \cline{4-9}
                            &                       & FB   & 35.93 $\pm$ 0.29 & 64.42 $\pm$ 0.56 & -16.79 $\pm$ 0.39 & 38.09  $\pm$ 0.35 & 91.53  & 49.31M \\
                            &                       & FB\_future    &	51.55 & 64.51 & -14.87 &	54.52 & 540.19 & 257.54M \\
                            \cline{2-9}
                            & \multirow{5}{*}{HyTE} & FT      & 39.01	 $\pm$ 0.53 & 33.49  $\pm$ 0.44 & \textbf{-11.62}  $\pm$ 0.20 & 43.00  $\pm$ 0.37 & 0.77 & 424.7K \\
                            &                       & TR      & 41.67	 $\pm$ 0.09 & \textbf{33.13}  $\pm$ 0.51 & -11.64  $\pm$ 0.14 & 46.15  $\pm$ 0.14 & 1.08 & 424.7K \\
                            &                       & TIE   & \textbf{42.41}  $\pm$ 0.24 & 35.02  $\pm$ 0.80 & -12.04  $\pm$ 0.21 & \textbf{47.00}  $\pm$ 0.01 & 9.63 & 1.41M \\
                            \cline{4-9}
                            &                       & FB     & 43.34  $\pm$ 0.16 & 68.34  $\pm$ 0.55 & -18.13  $\pm$ 0.38 & 46.14  $\pm$ 0.12 & 88.59 & 49.31M \\
                            % & 0.4246 & 0.3439 &	-0.1245	 &	0.4682 &	- &	- \\
                            &                       & FB\_future      & 56.88 & 69.27 & -15.38 &	60.43 & 922.53 & 257.54M \\                 
                            
    \midrule
    \multirow{10}{*}{YAGO11k} & \multirow{5}{*}{DE}    & FT   & 18.79	 $\pm$ 0.29 & 7.59	 $\pm$ 0.25 & -0.23	 $\pm$ 0.11 & 22.33  $\pm$ 0.19 & 0.84 & 296.2K	 \\
                            &                       & TR   & 20.82 $\pm$ 0.08 & 7.39  $\pm$ 0.03 & -0.12 $\pm$ 0.02  & 23.44 $\pm$ 0.02 & 1.19 & 296.2K \\
                            &                       & TIE  & \textbf{21.32}  $\pm$ 0.18 & \textbf{7.28}  $\pm$ 0.20 & \textbf{-0.05}  $\pm$ 0.04 & \textbf{23.95}  $\pm$ 0.23 & 11.27 & 1.28M \\
                            \cline{4-9}
                            &                       & FB   & 29.73  $\pm$ 0.06 & 57.61  $\pm$ 0.43 & -12.23  $\pm$ 0.03 & 31.23  $\pm$ 0.09 & 158.05 & 62.69M \\
                            &                       & FB\_future    & 33.26  & 32.90 & -6.37	& 34.93 & 300.98 &	215.89M \\
                            \cline{2-9}
                            & \multirow{5}{*}{HyTE} & FT       & 30.39  $\pm$ 0.10 & \textbf{36.68}  $\pm$ 0.50 & \textbf{-10.75}  $\pm$ 0.08 & 34.09  $\pm$ 0.07 & 0.59 & 296.2K \\
                            &                       & TR       & \textbf{31.99}  $\pm$ 0.24 & 38.93  $\pm$ 1.31 & -11.26  $\pm$ 0.50 & 35.80  $\pm$ 0.05 & 0.89 & 296.2K \\
                            &                       & TIE    & \textbf{31.99} $\pm$ 0.25 & 39.93 $\pm$ 0.90 & -11.63	 $\pm$ 0.41 & \textbf{35.91}  $\pm$ 0.18 & 8.82 & 1.28M \\
                            \cline{4-9}
                            &                       & FB  & 35.83	 $\pm$ 0.15 & 63.83	 $\pm$ 0.20 & -13.36 $\pm$ 0.07 & 37.43  $\pm$ 0.07 & 106.64 & 62.69M \\
                            &                       & FB\_future & 39.48 & 61.37 & -11.90	 & 41.29 & 663.36 & 215.89M \\     
  \bottomrule
\end{tabular}
  \caption{The overall performance comparison, averaged over four runs using different random seeds (except FB\_future). The mean and standard deviation results on test set are reported.}\label{tab:main_result}
  
\vspace*{-2ex}
\end{table*}

Table \ref{tab:main_result} reports the performances of the proposed methods and baselines. We make the following observations:
\begin{enumerate}[leftmargin=*]
    \item Our proposed method \textbf{TIE} achieves better C-Hits@10 and A-Hits@10 than both the FT and TR baselines (besides Wikidata12K + DE combination). It also performs better in terms of intransigence measures in DE experiments. 
    \item Although two full-batch skylines achieve better C-Hits@10 and A-Hits@10 than our proposed methods, they perform poorly on intransigence measures. This indicates the full-batch methods fail to identify obsolete facts promptly. Moreover, \textbf{FB\_future} is impractical in our setting because the model does not have access to future data. The practical skyline \textbf{FB} yields a relatively small advantage compared to the proposed methods on C-Hits@10 but performs dramatically worse on the intransigence measures.
    
    Despite the improvement, the RRD results of all methods are negative, meaning that on average, the deleted facts rank higher than the currently valid facts.  Further research is required to develop better strategies to reverse this.
    %Future exploration is required in this regard. 
    
    \item Comparing the different base models, HyTe outperforms DE in terms of C-Hits@10 and A-Hits@10 but under-performs for DF@10 and RRD. Moreover, augmenting the HyTE fine-tuning baseline with either TR or TIE does not result in an improvement. 
    % We hypothesize that it results from the lack of In HyTE, all the entities and relation embedding are projected using the same time embedding at each time step. 
    Both TR and experience replay emphasize remembering the previously valid facts and hence improve the A-Hits@10 metric significantly. However, due to the limited representation capacity of HyTE's time-embedding approaches, the model is unable further to distinguish the deleted facts from the previously true facts. This is partly because all the entity and relation embeddings are projected using the same time embedding at each time step. 
    
    \item \textbf{TIE} significantly improves both the time efficiency and data efficiency compared to full-batch training settings. The proposed  \textbf{TIE} sacrifices time efficiency compared to the TR method as the loss computation of the experience replay component is costly. Nevertheless, \textbf{TIE} and the \textbf{TR} methods reduce the training time by about 10 and 100 times, respectively comparing to \textbf{FB}.
\end{enumerate}

\subsection{Ablation Study}
To better understand the effectiveness of the proposed methods, we conduct the following four ablation experiments. We use DE as the base model on the Wikidata12k and YAGO11k datasets for ablation analysis. The test results are averaged over four randomized runs. Standard deviations are sometimes omitted due to space constraints. We use "Wiki" and "YAGO" as abbreviates of the two datasets. 

\subsubsection{Experience Replay Sampling Methods (\ref{positive_sampling})} As described in Sec \ref{positive_sampling}, we proposed multiple sampling strategies for sampling positive facts from the replay buffer. Table \ref{tab:ablation_5} shows the results of applying each sampling strategy, as well as no experience replay. We first observe that using positive facts leads to a better performance in almost all measures compared to not using experience replay, irrespective of the specific sampling strategy. 

Comparing among the different sampling strategies, the frequency-based sampling performs statistically better on C-Hits@10 and the intransigence measure than other alternatives for the Wikidata12k dataset. The inverse frequency-based sampling performs better on the C-Hits@10 and A-Hits@10 on average, while worse on the DF and RRD measures by average. However, the results for the YAGO11k data set are inconclusive. Considering the standard deviation, no method achieves a significant outperformance.

As the best sampling strategy in each combination of the base model and dataset varies, we treat the sampling strategy as a categorical hyperparameter and choose the one that yields the best validation performance for each experiment setting to apply on the test set. 

% In contrast to our intuition, the inverse frequency-based sampling does not result in a better intransigence measure. On the contrary, the inverse frequency-based sampling never outperforms the frequency-based sampling strategy in any setting. 
% Meanwhile, we observe that frequency-based sampling tends to achieve a better intransigence measure with comparable hits accuracy compared to uniform sampling. % Therefore, we use frequency-based sampling in the \textbf{TIE} model.

\begin{table}[ht]
% \vspace*{-2ex}
  
  \resizebox{\columnwidth}{!}{
      \begin{tabular}{c|cccccc}
        \toprule
        \multicolumn{1}{c}{Dataset} & Replay & C@10 ($\uparrow$) & DF@10 ($\downarrow$) & RRD ($\uparrow$) & A@10 ($\uparrow$) \\
        \midrule
        \midrule
        \multirow{4}{*}{Wiki} & None         & 33.33	$\pm$ 0.13 & 9.60 $\pm$ 0.23 & -2.57 $\pm$ 0.13 & 36.59 $\pm$ 0.05
 \\ 
                         & Uniform     & 34.70 $\pm$ 0.17	& 9.36 $\pm$ 0.50	& -2.12 $\pm$ 0.16	& \textbf{37.65} $\pm$ 0.15 \\
                         & Freq      & \textbf{34.90} $\pm$ 0.18	& \textbf{8.76} $\pm$ 0.66	& \textbf{-1.89} $\pm$ 0.36	& 37.57 $\pm$ 0.07 \\
                         & Inv-Freq     & 34.48 $\pm$ 0.12 & 9.37 $\pm$ 0.72 & -2.07 $\pm$ 0.43 & \textbf{37.62} $\pm$ 0.12 \\
        \midrule
        \multirow{4}{*}{YAGO}    & None         & 20.65 $\pm$ 0.23	& \textbf{7.18} $\pm$ 0.14	& \textbf{-0.03} $\pm$ 0.03	& 23.25 $\pm$ 0.18 \\ 
                        & Uniform     & 21.07 $\pm$ 0.15	& \textbf{7.19} $\pm$ 0.14	& -0.13 $\pm$ 0.03	& 23.70 $\pm$ 0.12 \\
                        & Freq     & 21.29 $\pm$ 0.15	& 7.22 $\pm$ 0.11	& -0.13 $\pm$ 0.07	& 23.85 $\pm$ 0.19 \\
                        & Inv-Freq     & \textbf{21.32} $\pm$ 0.18	& 7.28 $\pm$ 0.20	& \textbf{-0.05} $\pm$ 0.04	& \textbf{23.95} $\pm$ 0.20 \\
      \bottomrule
      \end{tabular}
    }
  \caption{Ablation analysis on different experience replay strategies. All experiments use added facts, deleted facts and temporal regularization. Uniform, Freq, and Inv-Freq stand for uniform sampling, frequency-based sampling, and inverse frequency based sampling respectively.}
  \label{tab:ablation_5}
  \vspace{-2ex}
\end{table}
\subsubsection{Learning with Deleted Facts (\ref{sec:deleted_facts_sampling})} To evaluate the effectiveness of the deleted fact sampling component, we compare the performance of adding the deleted facts (Del) to the following baseline models: 1) temporal regularization (TR); 2) the combination of temporal regularization and experience replay (TR + Replay). All models use added facts by default.  As shown in Table \ref{tab:ablation_1}, adding sampled deleted facts consistently improves the model's intransigence measure in DF-Hits@10 and RRD. Meanwhile, it slightly impedes the model's performance in terms of C-Hits@10 and A-Hits@10 for most cases. This demonstrates that the deleted fact sampling helps the model differentiate triples with low ranks from triples with high ranks with a limited sacrifice in the model's ranking performance. Practitioners should be mindful of this trade-off when deciding whether to use deleted facts.

\begin{table}[ht]
  \resizebox{\columnwidth}{!}{
      \begin{tabular}{c|cccccccc}
        \toprule
        \multicolumn{1}{c}{Dataset} & TR & Replay & Del & C@10 ($\uparrow$) & DF@10 ($\downarrow$) & RRD ($\uparrow$) & A@10 ($\uparrow$) \\
        \midrule
        \midrule
        \multirow{4}{*}{Wiki} & \checkmark & &               & \textbf{34.14}	& 13.05	& -4.45	& \textbf{38.43} \\
                              & \checkmark & & \checkmark           & 33.33	& \textbf{9.60}	& \textbf{-2.57}	& 36.60 \\
                            \cline{2-8}
                              & \checkmark & \checkmark &  &	\textbf{3614}	& 15.07 & -0.45	& \textbf{39.87} \\
                              & \checkmark &\checkmark &\checkmark   & 34.90	& \textbf{8.76} & \textbf{-1.89}	& 37.57 \\
        \midrule
        \multirow{4}{*}{YAGO} & \checkmark & &   & \textbf{20.82}	& 7.39	& -0.12	& \textbf{23.44} \\
                              & \checkmark & & \checkmark & 20.65	& \textbf{7.18}	& \textbf{-0.03}	& 23.25 \\
                            \cline{2-8}
                              & \checkmark & \checkmark &  & \textbf{22.09}	& 7.77	& -0.25	& \textbf{24.29} \\
                               & \checkmark & \checkmark &\checkmark & 21.32 & \textbf{7.28}	& \textbf{-0.05}	& 23.95 \\
      \bottomrule
      \end{tabular}
    }
  \caption{Ablation analysis of applying deleted facts sampling. Replay and Del stand for the experience replay and deleted facts training respectively.}
  \vspace{-2ex}
  \label{tab:ablation_1}
\end{table}

\subsubsection{Learning with Added Facts (\ref{sec:added_facts})} To evaluate the impact of only using newly added facts for incremental training, we conduct experiments to compare 1) only using added facts $D^t_{add}$ and 2) using all the new facts $D^t$ for training at each time step. Table \ref{tab:ablation_2} shows that training with all facts generally yields slightly better or comparable performance across all measures to training with only added facts. More specifically, training with all facts consistently performs better on C-Hits@10. 
%as it tends to overfit to the current time step. 
However, there exists a trade-off between sampling efficiency and other metrics. The sizes of training data required by added-facts experiments are only 9\% and 4.62\% of those required by the all-facts experiments for Wikidata12k and YAGO11k, respectively. 

% Taking TR methods as an example, training with only added facts reduces the training time by nearly 10 times while yielding comparable results to training with all facts. Under circumstances where the model needs to be frequently updated, it is favorable to trade a minor loss in ranking performance to save the training time.

\begin{table}[ht]
  \resizebox{\columnwidth}{!}{
      \begin{tabular}{c|ccccc}
        \toprule
        \multicolumn{1}{c}{Dataset} & Algo. & C@10 ($\uparrow$) & RRD ($\uparrow$) & A@10 ($\uparrow$) & Data size  \\
        \midrule
        \midrule
        \multirow{4}{*}{Wiki} & FT-Add    & 31.90 $\pm$ 0.42 & -4.16 $\pm$ 0.07 & \textbf{36.05} $\pm$ 0.52 & \textbf{424.7K} \\
                              & FT-All    & \textbf{33.54} $\pm$	0.39 & \textbf{-4.13} $\pm$	0.55 & 35.11 $\pm$ 0.36 & 4.71M\\
                              \cline{2-6}
                              & TR-Add    & 34.14 $\pm$	0.17 & \textbf{-4.45} $\pm$	0.09 & 38.43 $\pm$ 0.13 & \textbf{424.7K} \\
                              & TR-All    & \textbf{34.92} $\pm$	0.20 & -4.63 $\pm$	0.12 & \textbf{38.62} $\pm$ 0.08 & 4.71M \\
        \midrule
        \multirow{4}{*}{YAGO} & FT-Add          & 18.79 $\pm$ 0.29	& \textbf{-0.23} $\pm$ 0.11	& 22.33 $\pm$ 0.34 & \textbf{296.2K} \\
                              & FT-All          & \textbf{23.45} $\pm$ 0.22	& -2.73 $\pm$ 2.18	& \textbf{25.07} $\pm$ 0.27 & 6.40M \\
                              \cline{2-6}
                              & TR-Add          & 20.82 $\pm$ 0.08	& -0.12 $\pm$ 0.02	& 23.44 $\pm$ 0.03 & \textbf{296.2K} \\
                              & TR-All          & \textbf{21.50} $\pm$ 0.05	& \textbf{-0.00} $\pm$ 0.02  & \textbf{23.91} $\pm$ 0.10  & 6.40M \\
      \bottomrule
      \end{tabular}
    }
  \caption{Ablation analysis of only using added facts for incremental learning. "Add" in the Algo. column means the model uses only added facts at each incremental training step. In contrast, "All" means the model uses all facts.}
    \vspace*{-3ex}
  \label{tab:ablation_2}
\end{table}

\subsubsection{Experience Replay Sampling Size(\ref{sec:experience_replay})} 
To understand how the size of replay data affects the model's performance, we conduct experiments altering the size of the sampled replay facts on Wiki dataset. For Table \ref{tab:ablation_3}, we fix the window length $\tau = 10$ and vary the replay sampling size for each step. For Table \ref{tab:ablation_4}, we vary $\tau$ and the sample size jointly. %, where the sample size grows linearly with the $\tau$.

Table \ref{tab:ablation_3} shows all measures tend to improve as the sampling size increases while keeping $\tau$ fixed. 
%The largest sampling size of 40,000 achieves the best result for all measures except DF-Hits@10. 
On the contrary, Table \ref{tab:ablation_4} shows that larger replay data size does not always lead to better performance as the model's performance is sensitive to the value of $\tau$, which is the window length of the old data. Among experiments from Table \ref{tab:ablation_4}, best performance is achieved when $\tau$ and replay size are 15 and 15,000, respectively. Meanwhile, from both tables we notice that training time increases as the replay sample size grows. Performance and training time trade-off needs to be considered while choosing hyper-parameters.

% The best C-Hits@10, A-Hits@10, and DF-Hits@10 are reached when sampling 15,000 positive replay facts with the time window size of 15. Further increasing the sampling size beyond 20,000 or time window size beyond 20 leads to a decrease in all measures. We conclude that a larger sampling size generally leads to better performance. Meanwhile, replaying the historical facts within a certain time range benefits the model, and the effectiveness of the experience replay component is sensitive to the time window size.

\begin{table}[ht]
  \resizebox{\columnwidth}{!}{
      \begin{tabular}{cc|cccccc}
        \toprule
        sample size & $\tau$ & C@10 ($\uparrow$) & DF@10 ($\downarrow$) & RRD ($\uparrow$) & A@10 ($\uparrow$) & Training time \\
        \midrule
        \midrule
        1000	&	10	&	36.17	        &	11.60	        &	-3.04	            &	38.84	        &	6.07 \\
        3000	&	10	&	36.02	        &	10.07	        &	-2.09	            &	38.64	        &	8.02 \\
        10000	&	10	& 34.90 & \textbf{8.76} & -1.89 	& 37.57 & 15.79 \\
        20000	&	10	&	36.71	        &	9.08	&	-1.72	            &	39.32	        &	20.30 \\
        40000	&	10	&   \textbf{37.21}	&	9.34	        &	\textbf{-1.59}	&	\textbf{39.53}	&	31.83 \\
      \bottomrule
      \end{tabular}
    }
  \caption{Ablation analysis of using different experience replay sampling size.}
  \vspace*{-3ex}
  \label{tab:ablation_3}
\end{table}

\begin{table}[ht]
  \resizebox{\columnwidth}{!}{
      \begin{tabular}{cc|cccccc}
        \toprule
        sample size & $\tau$ & C@10 ($\uparrow$) & DF@10 ($\downarrow$) & RRD ($\uparrow$) & A@10 ($\uparrow$) & Training time \\
        \midrule
        \midrule
        5000	&	5	&	36.87	        &	11.09	        &	-2.45	            &	39.43	        &	8.79  \\
        10000	&	10	& 34.90 & \textbf{8.76} & \textbf{-1.89} 	& 37.57 & 15.79 \\
        15000	&	15	&	\textbf{36.94}	&	9.65	&	-2.24	            &	\textbf{39.53}	&	19.61 \\
        20000	&	20	&	36.36	        &	9.80	        &	-1.94	&	35.02	        &	25.95 \\
        25000	&	25	&	36.41	        &	10.43	        &	-2.22	            &	35.12      	&	28.92 \\
      \bottomrule
      \end{tabular}
    }
  \caption{Ablation analysis on different experience replay data size.}
  \label{tab:ablation_4}
\vspace*{-6ex}
\end{table}
% \subfile{fine_grain_analysis}

\section{Conclusion}
We present a novel incremental learning framework named TIE for TKGC tasks. TIE combines TKG representation learning, frequency-based experience replay, and temporal regularization to improve the model's performance on both current and past time steps. 
TIE leverages pattern frequencies to select among reservoir samples and uses only the deleted and added facts at the current time step for training, which significantly reduces training time and the size of training data.  Moreover, we propose DF and RRD metrics to measure the intransigence of the model.
Extensive ablation studies shows each proposed component's effectiveness. They also provide insights for deciding among model variations by revealing performance trade-offs among various evaluation metrics.

This work serves as a first attempt and exploration to apply incremental learning to TKGC tasks.
Future work might involve exploring other incremental learning techniques, such as constrained optimization, to achieve more robust performance across datasets and metrics.

\section*{Acknowledgement}
This research was supported in part by Noah’s Ark Lab (Montreal Research Centre), CIFAR Canada AI Chair program, FRQNT \footnote{The Fonds de Nature et technologies of Quebec} and Samsung Electronics. The authors would like to thank Noah’s Ark Lab for providing the computational resources.
\bibliographystyle{ACM-Reference-Format}
\balance
\bibliography{corrected}

\appendix
\newpage
\section{Appendix}
\subsection{A-GEM Formulation and Adaptation}\label{appendix:a-gem}
\begin{figure*}
  \includegraphics[width=0.8\linewidth]{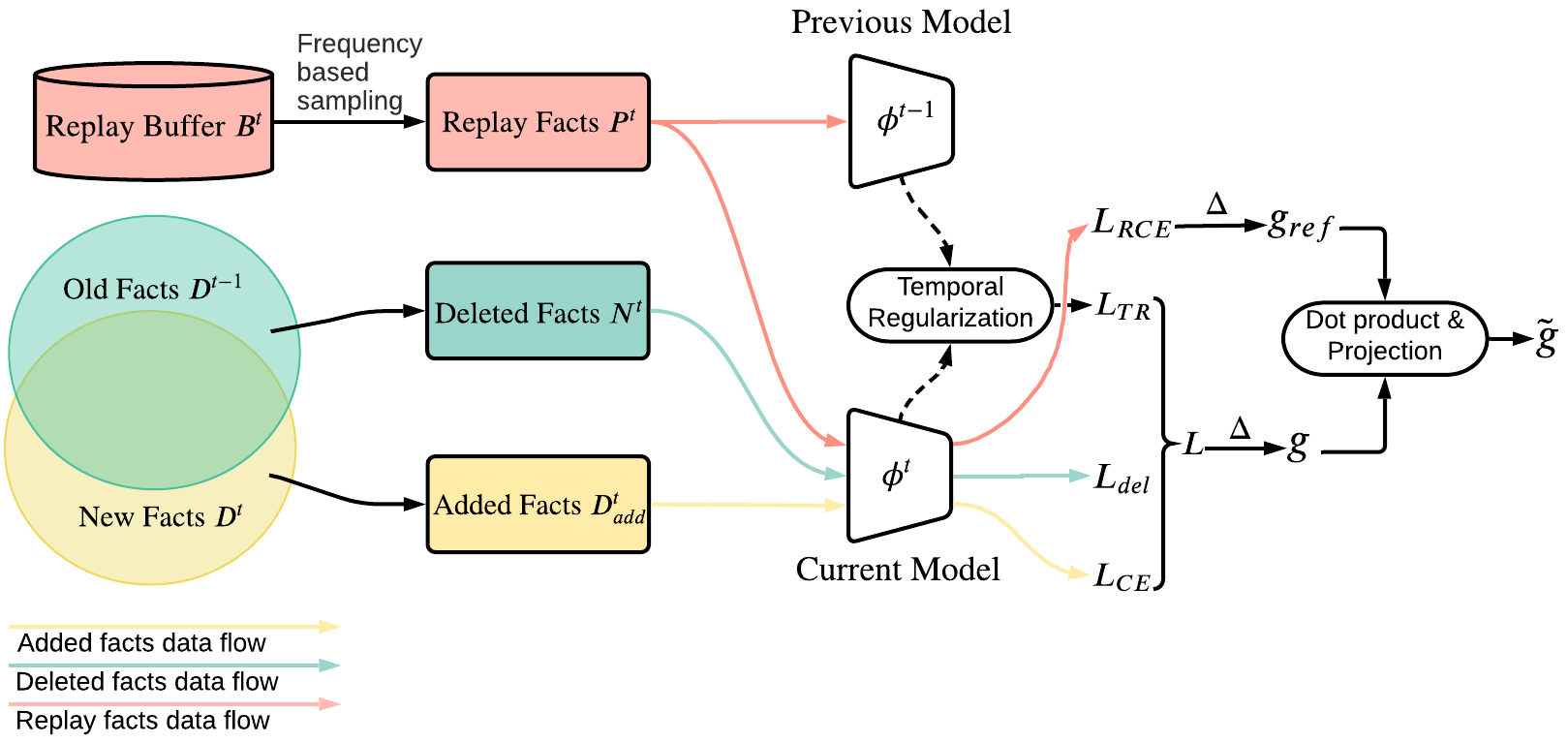}
    \caption{Architecture of the A-GEM variation of TIE model. }
\label{fig:a-gem}
\end{figure*}

In this section, we introduce the A-GEM formulation and the its adaptation to our task in detail.  Figure \ref{fig:a-gem} presents the architecture of A-GEM variation in details. 

\paragraph{Formulation}
The goal of A-GEM~\cite{chaudhry2018efficient} is to optimize the loss on the current samples without increasing the loss on the previous samples, which is approximated by the \emph{average} loss. At every time step, A-GEM insures that the average loss on the reservoir samples does not increase. We slightly modify the original formulation to adapt to the TKGC task. At time step $t$, the objective of A-GEM is:
\begin{equation}\label{a-gem-original}
    \text{minimize} \quad L(\phi^t, D^t) \quad \text{s.t.} \quad L(\phi^t, P^t) \leq L(\phi^{t-1}, P^t).
\end{equation}

Replacing the loss terms using our proposed loss terms in the previous section, the above objective is written as:

\begin{equation}\label{a-gem}
    \text{minimize} \quad L_{CE} + L_{del} + L_{TR} \quad \text{s.t.} \quad L^t_{RCE} \leq L^{t-1}_{RCE},
\end{equation}
where $L^t_{RCE}$ is the same loss term as in equation \ref{equation:rce}, and $L^{t-1}_{RCE}$ is written as:
\begin{equation}
    L^{t-1}_{RCE} = -\sum_{s, r, o, t' \in P^t} log(q^{t-1}_{s, r, o, t'}).
\end{equation}

Let $g$ denote the gradient update on the current loss, i.e. $L_{CE} + L_{del} + L_{TR}$, and $g_{ref}$ denote the gradient of $L^t_{RCE}$. The above optimization problem is shown to be equivalent to:
\begin{equation}
    minimize_{\tilde{g}} \frac{1}{2} ||g - \tilde{g}||^2 \quad \textrm{s.t.} \quad \tilde{g}^T g_{ref} \geq 0.
\end{equation}
When violating the constraints, i.e. $\tilde{g}^T g_{ref} < 0$, the solution is derived as: 
\begin{equation}
    \tilde{g} = g - \frac{g^T g_{ref}}{g_{ref}^T g_{ref}} g_{ref}.
\end{equation}
\paragraph{Modified A-GEM parameter update} A-GEM was developed and evaluated on image classification tasks, where parameters updates are performed on the neural network parameters. When optimizing the embedding based models, however, problems arise. Suppose the entity \textit{united states} is involved in some facts in $P^t$ but can't be found in $D^t$. In the standard optimization setting, the gradient of the embedding of \textit{united states} will be applied. However, in A-GEM, if the constraint $g^T g^t_{ref} \geq 0$ is violated, which is computed using gradients of all parameters in the model, then the embedding of all entities likes \textit{united states} will not be updated. This will incur huge cost of gradient computation and less optimal parameter optimization.

To address this issue, we shift the target of the constraint from global gradient matrix to each vector representation. Let $g_i$ and ${g_{ref}}_i$ denote the i-th row of the gradient matrices $g$ and $g_{ref}$ respectively. The optimization goal is modified to:
\begin{equation}
    minimize_{\tilde{g}_i} \frac{1}{2} ||g_i - \tilde{g}_i||^2 \quad \textrm{s.t.} \quad \tilde{g}_i^T {g_{ref}}_i \geq 0, \forall i. 
\end{equation}
The projected gradient matrix $\tilde{g}$ is:

\begin{equation}
     \tilde{g}_i =
    \begin{cases}
         g_i, & \text{if } \tilde{g}_i^T {g_{ref}}_i \geq 0, \\
         g_i - \frac{g_i^T {g_{ref}}_i}{{g_{ref}}_i^T {g_{ref}}_i} {g_{ref}}_i, & \text{otherwise}.
    \end{cases}
\end{equation}
\subsection{Supplementary Ablation Study}\label{app:ablation}
\subsubsection{Experience replay buffer size(\ref{sec:experience_replay})} To examine the effectiveness of the experience replay component, we conduct experiments by altering the maximum number of the positive replay facts sampled from the replay buffer. We alter the replay sample sizes in two separate experiments. First, in Table \ref{tab:ablation_3}, we fix the window length $\tau = 10$ and vary the replay sampling size. Second, in Table \ref{tab:ablation_4}, we vary both $\tau$ and the sample size, where the sample size grows linearly with the $\tau$.

As observed from Table \ref{tab:ablation_3}, all measures tend to improve as the sampling size increases. The largest sampling size of 40,000 achieves the best result for all measures except DF-Hits@10. At the same time, we notice that each time step's training time also increases as the replay sample size grows. There exists a trade-off between performance and training time.

Interestingly, unlike the result from table \ref{tab:ablation_3}, we observe in table \ref{tab:ablation_4} that larger data size does not lead to a better performance in terms of neither Hits@10 measure nor intransigence measure. The best C-Hits@10, A-Hits@10, and DF-Hits@10 are reached when sampling 15,000 positive replay facts with the time window size of 15. Further increasing the sampling size beyond 20,000 and time window size beyond 20 leads to a decrease in all measures. Combining with the previous observation, we conclude that a larger sampling size generally leads to better performance. Meanwhile, replaying the historical facts within a certain time range benefits the model, and the effectiveness of the experience replay component is sensitive to the time window size.

\subsection{A-GEM optimization}

We report the comparative study between the original TIE model and the A-GEM variation of TIE model in Table \ref{tab:ablation_6}. The difference between the model performance with and without A-GEM are not statistically significant. This echos the finding in \cite{chaudhry2019continual} that simple experience replay methods are empirically better than A-GEM on incremental image classification tasks. Our result can be seen an extended piece of evidence in the domain of incremental knowledge graph completion.

\begin{table*}[!htb]
%   \resizebox{\columnwidth}{!}{
      \begin{tabular}{c|ccccccc}
        \toprule
        Dataset & Replay & A-GEM & C@10 ($\uparrow$) & DF@10 ($\downarrow$) & RRD ($\uparrow$) & A@10 ($\uparrow$) \\
        \midrule
        \midrule
        \multirow{6}{*}{Wiki} & Uniform &             & 34.70 $\pm$ 0.17	& 9.36 $\pm$ 0.50 & \textbf{-2.12} $\pm$ 0.16	& \textbf{37.65} $\pm$ 0.15 \\
                              & Uniform &\checkmark   & \textbf{34.84} $\pm$ 0.16	& \textbf{9.23} $\pm$ 0.30 & -2.22 $\pm$ 0.23	& 37.66 $\pm$ 0.15 \\
                            \cline{2-7}
                              & Freq &                       & \textbf{34.90} $\pm$ \textbf{0.18}	& \textbf{8.76} $\pm$ 0.66	& \textbf{-1.89} $\pm$ 0.36	& \textbf{37.57} $\pm$ 0.07 \\
                               & Freq & \checkmark           & 34.76 $\pm$ 0.20	& 8.87 $\pm$ 0.59   & -1.98 $\pm$ 0.25	& 37.45 $\pm$ 0.11 \\
                            \cline{2-7}
                              & Inv-Freq &              & 34.48 $\pm$ 0.12 & \textbf{9.37} $\pm$ 0.72 & \textbf{-2.07} $\pm$ 0.43 &\textbf{37.62} $\pm$ 0.12 \\
                              & Inv-Freq &\checkmark    & 34.47 $\pm$ 0.35 & 9.61 $\pm$ 0.26 & -2.29 $\pm$ 0.20 & 37.47 $\pm$ 0.18\\
        \midrule
        \multirow{6}{*}{YAGO}  & Uniform  &          & 21.07 $\pm$ 0.15	& \textbf{7.19} $\pm$ 0.14	& \textbf{-0.13} $\pm$ 0.03	& 23.70 $\pm$ 0.14 \\
                               & Uniform &\checkmark & \textbf{21.55} $\pm$ 0.20	& 7.35 $\pm$ 0.15	& -0.26 $\pm$ 0.10	& \textbf{23.97} $\pm$ 0.12 \\
                            \cline{2-7}
                               & Freq   &           & \textbf{21.29} $\pm$ 0.15	& 7.22 $\pm$ 0.11	& -0.13 $\pm$ 0.07	& \textbf{23.85} $\pm$ 0.16 \\
                               & Freq & \checkmark  & 21.21 $\pm$ 0.24	& \textbf{7.16} $\pm$ 0.12	& -0.15 $\pm$ 0.06	& 23.76 $\pm$ 0.19 \\
                            \cline{2-7}
                              & Inv-Freq  &             & 21.32 $\pm$ 0.18	& \textbf{7.28} $\pm$ 0.20	& \textbf{-0.05} $\pm$ 0.04	& \textbf{23.95} $\pm$ 0.23 \\
                              & Inv-Freq &\checkmark    & 21.31 $\pm$ 0.36	& 7.35 $\pm$ 0.18	& -0.17 $\pm$ 0.12	& 23.87 $\pm$ 0.20 \\
      \bottomrule
      \end{tabular}
    % }
  \caption{Ablation analysis of A-GEM under using sampling methods. Replay stands for the type of replay sampling method and A-GEM column indicates whether A-GEM is used or not. All experiments use added facts, deleted facts and temporal regularization bu default. }
  \label{tab:ablation_6}
\end{table*}

\end{document}